\documentclass{article}
\usepackage{graphicx} 
\usepackage[margin=1in]{geometry}
\usepackage{authblk}
\usepackage[numbers]{natbib}
\usepackage[colorlinks=true,linkcolor=black,citecolor=blue,urlcolor=blue,]{hyperref}
\usepackage{amsmath}
\usepackage{algorithm}
\usepackage{algorithmic}
\usepackage{multirow}
\usepackage{array}
\usepackage{longtable}
\usepackage[final]{changes}
\usepackage{lineno}
\usepackage{comment}

\title{Automated Measurement of Optic Nerve Sheath Diameter Using Ocular Ultrasound Video}
\author[1,*]{Renxing Li}
\author[2,*]{Weiyi Tang}
\author[3,*]{Peiqi Li}
\author[4]{Qiming Huang}
\author[1,4]{Jiayuan She}
\author[3]{Shengkai Li}
\author[6]{Haoran Xu}
\author[8]{Yeyun Wan}
\author[5]{Jing Liu}
\author[7,$\dag$]{Hailong Fu}
\author[2,$\dag$]{Xiang Li}
\author[1,$\dag$]{Jiangang Chen}

\affil[1]{\small Shanghai Key Laboratory of Multidimensional Information Processing, East China Normal University, Shanghai, 200241, China \\
	\texttt{aa1773894229@163.com (Renxing Li); Lucas\_sjy@outlook.com (Jiayuan She); jgchen@cee.ecnu.edu.cn (Jiangang Chen)}}

\affil[2]{Intensive Care Unit, Minhang Hospital, Fudan University, Shanghai, 201199, China \newline
	\texttt{20211360008@fudan.edu.cn (Weiyi Tang); 18918169826@163.com (Xiang Li)}}

\affil[3]{School of Mathematics and Physics, Xi'an Jiaotong-Liverpool University, Suzhou, Jiangsu, 215123, China \\
	\texttt{LPQ\_0619@outlook.com (Peiqi Li); Shengkai.Li23@student.xjtlu.edu.cn (Shengkai Li)}}

\affil[4]{School of Artificial Intelligence and Advanced Computing, Xi'an Jiaotong-Liverpool University (Taicang Campus), Taicang, Jiangsu, 215400, China \\
	\texttt{thomaszz4huang@gmail.com (Qiming Huang);}}

\affil[5]{Department of Neonatology and NICU, Beijing Obstetrics and Gynecology Hospital, Capital Medical University, Beijing, 100026, China \newline
	\texttt{liujingbj@live.cn (Jing Liu)}}

\affil[6]{Khoury College of Computer Science, Northeastern University, Boston, MA, 02115, United States of America \newline
	\texttt{xu.haoran3@northeastern.edu (Haoran Xu)}}

\affil[7]{Department of Anesthesiology, Changzheng Hospital, Second Affiliated Hospital of Naval Medical University, Shanghai, 200003, China \newline
	\texttt{fuhailong1979@163.com (Hailong Fu)}}

\affil[8]{School of Advanced Technology, Xi’an Jiaotong-Liverpool University, Suzhou, Jiangsu, 215123, China \newline 
	\texttt{Yeyun.Wan22@student.xjtlu.edu.cn (Yeyun Wan)}}

\affil[*]{Renxing Li, Weiyi Wang, and Peiqi Li are the co-first authors.}
\affil[$\dag$]{Jiangang Chen, Xiang Li, and Hailong Fu are the corresponding authors.}

\date{}

\begin{document}
	
	\maketitle
	
	\begin{abstract}
		\textit{Objective}. Elevated intracranial pressure (ICP) is recognized as a biomarker of secondary brain injury, with a significant linear correlation observed between optic nerve sheath diameter (ONSD) and ICP. Frequent monitoring of ONSD could effectively support dynamic evaluation of ICP. However, ONSD measurement is heavily reliant on the operator's experience and skill, particularly in manually selecting the optimal frame from ultrasound sequences and measuring ONSD. \textit{Approach}. This paper presents a novel method to automatically identify the optimal frame from video sequences for ONSD measurement by employing the Kernel Correlation Filter (KCF) tracking algorithm and Simple Linear Iterative Clustering (SLIC) segmentation algorithm. The optic nerve sheath is mapped and measured using a Gaussian Mixture Model (GMM) combined with a KL-divergence-based method. \textit{Results}. When compared with the average measurements of two expert clinicians, the proposed method achieved a mean error, mean squared deviation, and intraclass correlation coefficient (ICC) of 0.04, 0.054, and 0.782, respectively. \textit{Significance}. The findings suggest that this method provides highly accurate automated ONSD measurements, showing potential for clinical application.
	\end{abstract}
	
	\textbf{Key words}: Optic Nerve Sheath Diameter; Intracranial Pressure; Ultrasound Video Analysis; Automated Frame Selection
	
	\section{Introduction}
	\label{sec:intro}
	Increased Intracranial \replaced{Pressure}{pressure} (ICP) is a critical biomarker which signals the presence of secondary brain injuries\cite{ussahgij2020optic, wang2019ultrasonographic}, which can result from various neurological conditions, including traumatic brain injuries (TBI), cerebral venous sinus thrombosis, cerebral hemorrhage, hydrocephalus, or intracranial inflammation\cite{smith2008monitoring,carney2017guidelines}. Accurate and timely monitoring of ICP is crucial, as elevated ICP is associated with severe complications such as irreversible ischemic and hypoxic encephalopathy, cerebral herniation, and even death\cite{williams2017optic,netteland2023noninvasive}. Thus, the early detection and management of elevated ICP are essential to improving outcomes in patients with acute brain injuries.
	
	ICP monitoring techniques are broadly classified into invasive and non-invasive methods. The extra-ventricular drain (EVD) technique, considered the gold standard for ICP monitoring, provides accurate real-time data\cite{raboel2012intracranial, jenjitranant2020correlation} but is associated with significant risks, including hemorrhage, infection, and catheter displacement\cite{nag2019intracranial,xu2022non}. These risks have spurred the development of non-invasive ICP assessment methods, which offer a safer alternative while providing clinically relevant information\cite{zweifel2012reliability,schmidt2000evaluation}. Among these methods, the measurement of optic nerve sheath diameter (ONSD) via ocular ultrasound has garnered considerable attention due to its non-invasive nature, ease of use, and ability to deliver rapid bedside assessments. The ONSD is a well-established surrogate marker for ICP, with numerous studies confirming a significant linear correlation between ONSD and ICP\cite{jalayondeja2021correlations}.
	
	Despite its clinical utility, the traditional methods for measuring ONSD are fraught with challenges\cite{montorfano2021mean}. The current practice of ONSD measurement is highly dependent on the operator's expertise, particularly in manually selecting the optimal frame from ultrasound sequences and accurately measuring the ONSD. This dependence introduces variability and subjectivity into the measurements, which can lead to inconsistencies in clinical decision-making. Moreover, the manual nature of these measurements makes them time-consuming and prone to inter-operator variability, limiting their effectiveness in emergency settings where rapid and reliable assessment is crucial.
	
	To address these challenges, recent research has focused on developing automated and semi-automated methods for ONSD measurement that reduce operator dependency and improve measurement accuracy. For instance, several studies have employed machine learning and computer vision techniques to enhance the reliability of ONSD measurements. These methods include the use of convolutional neural networks (CNNs) for automated image segmentation and classification, as well as advanced algorithms for frame selection and optic nerve sheath localization. While these approaches have shown promise, they often rely on complex model architectures and require large annotated datasets for training, which may not be readily available in all clinical settings.
	
	In response to these limitations, this study proposes a novel automated method for ONSD measurement that leverages the strengths of traditional image processing techniques while minimizing the need for extensive training data. Specifically, we utilize a Kernel Correlation Filter (KCF) tracking algorithm combined with Simple Linear Iterative Clustering (SLIC) segmentation to automatically identify the optimal frame from ocular ultrasound video sequences\cite{copetti2009optic,chen2019ultrasonic,maissan2015ultrasonographic}. Following frame selection, the optic nerve sheath is accurately mapped and measured using a Gaussian Mixture Model (GMM) and a KL-divergence-based approach. This method not only automates the ONSD measurement process but also enhances its robustness and accuracy, making it more suitable for clinical application, particularly in settings where rapid assessment is necessary.
	
	The proposed method is designed to address the key challenges of existing approaches, including the reliance on manual frame selection and the need for precise boundary detection in low-quality ultrasound images. By integrating Kernel Correlation Filter (KCF) tracking and SLIC segmentation, our method provides a more consistent and reliable framework for ONSD measurement, reducing inter-operator variability and improving diagnostic consistency . Furthermore, the use of a GMM with KL-divergence allows for more accurate boundary refinement, even in the presence of noise and imaging artifacts, further enhancing the method's clinical applicability.
	
	This study aims to contribute to the ongoing development of non-invasive ICP monitoring techniques by providing a robust, automated solution for ONSD measurement that can be easily implemented in clinical practice. We hypothesize that the proposed method will demonstrate superior accuracy and reliability compared to traditional manual methods, thereby supporting more timely and informed clinical decision-making in patients with suspected elevated ICP. In the following sections, we detail the methodology, experimental validation, and potential clinical implications of our approach.
	
	\section{Related Work}
	\label{sec:review}
	The measurement of optic nerve sheath diameter (ONSD) has become a crucial non-invasive technique for assessing intracranial pressure (ICP) in clinical practice. As the relationship between ONSD and ICP has been well-documented, numerous studies have focused on improving the accuracy, reliability, and accessibility of ONSD measurements. Traditional methods often involve manual frame selection from ultrasound sequences, which is not only time-consuming but also highly dependent on the operator's expertise, leading to significant inter- and intra-operator variability. This section explores various approaches that have been developed to automate ONSD measurement, focusing on their advantages, limitations, and the current state-of-the-art.
	
	\subsection{Traditional and Manual Methods}
	The manual measurement of optic nerve sheath diameter (ONSD) using B-mode ocular ultrasound has long been a cornerstone in the assessment of intracranial pressure and other neurological conditions. This method relies heavily on the skill and experience of clinicians to accurately capture and interpret ultrasound images. Despite its widespread use, manual measurement is inherently subject to operator variability, which can impact the consistency and reliability of the results. As such, it requires significant expertise to minimize errors and ensure reproducibility\cite{rajajee2021novel}.
	
	Manual techniques typically involve identifying the optic nerve sheath on a transverse plane, approximately 3 mm behind the globe, and measuring its diameter. This approach, while effective, is time-intensive and demands careful attention to image quality and patient positioning. The accuracy of manual ONSD measurement has been validated in numerous clinical settings, reinforcing its status as a reliable method for diagnosing conditions like intracranial hypertension. However, the reliance on operator skill highlights the need for standardized training and protocols to reduce inter-observer variability\cite{pang2019measurement}.
	
	Furthermore, as automated techniques begin to emerge, comparisons with manual methods have underscored the continuing importance of high standards in manual measurement practices. These studies emphasize that while automation may offer consistency, the manual measurement of ONSD remains a valuable diagnostic tool, particularly in environments where advanced automated systems may not be accessible\cite{romagnosi2020eyeing}.
	
	\begin{figure}[t]
		\centering
		\includegraphics[width=1.0\linewidth]{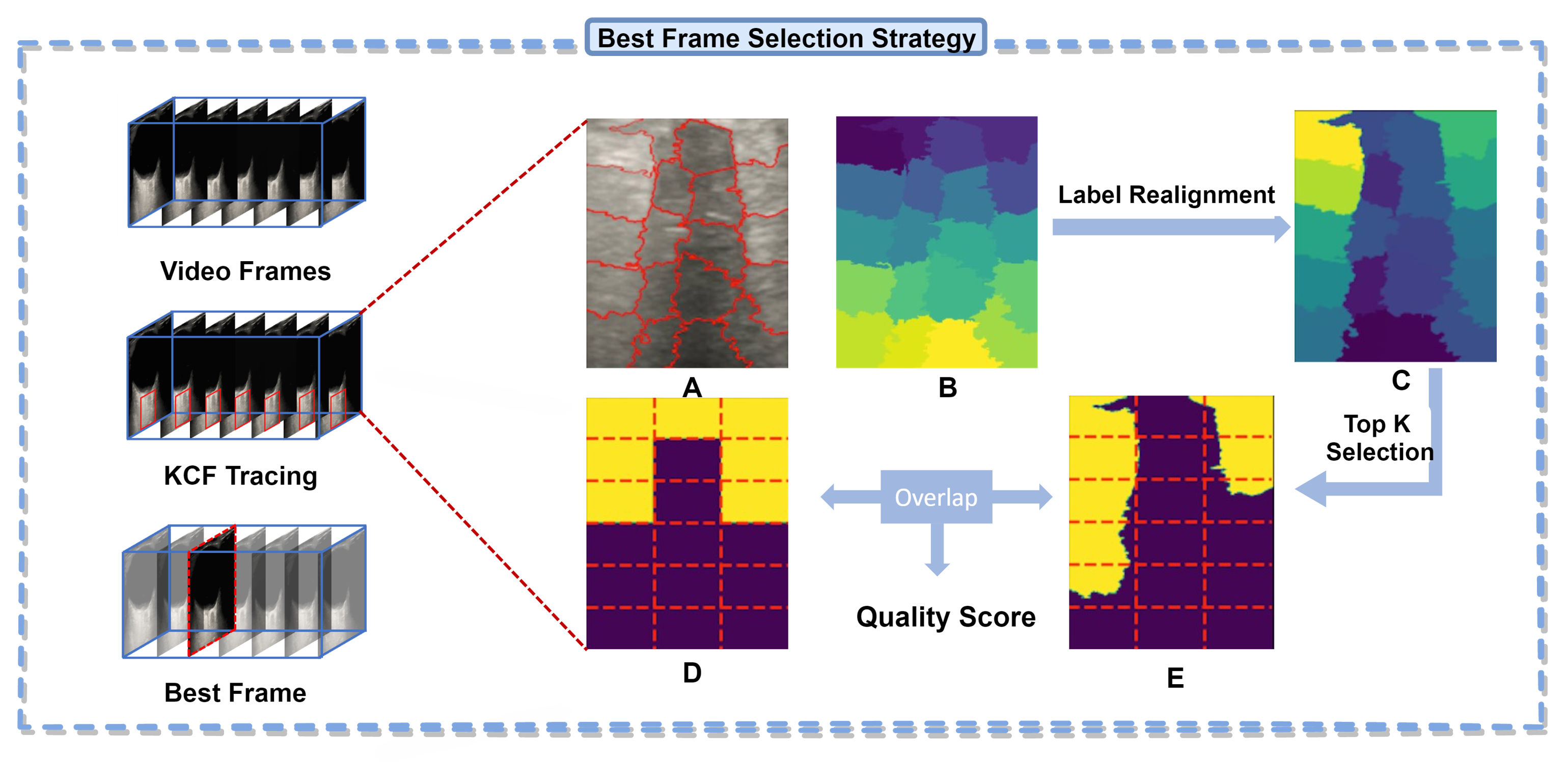}
		\caption{\textbf{Overall structure of optimal frame selection strategy}: (A) SLIC super-pixel segmentation edge result; (B) Predicted labels given by SLIC; (C) Result after label realignment; (D) Manually designed ground truth representing a good quality image cut of optic nerve sheath; (E) Segmentation of Top-K selection result.}
		\label{fig:1}
	\end{figure}
	
	\subsection{Semi-automated Methods}
	Recent advancements in the measurement of optic nerve sheath diameter (ONSD) have led to the development of semi-automated methods that aim to combine the accuracy of automated systems with the flexibility of manual techniques. One notable development is the use of a semi-automated approach to measure the ONSD to eyeball transverse diameter ratio (OER), which has demonstrated a stronger correlation with raised intracranial pressure than traditional ONSD measurements alone. This method not only enhances diagnostic accuracy but also reduces the time and variability associated with purely manual measurements, making it a valuable tool in clinical settings where rapid and precise assessment is critical\cite{singh2024novel}.
	
	Moreover, evidence suggests that semi-automated methods significantly improve measurement consistency and reduce inter-observer variability. Comparative studies between manual and semi-automated techniques in various imaging contexts, such as spine measurements and carotid intima-media thickness, have consistently shown that semi-automated methods offer superior intra-observer agreement and expedited processing times. These findings underscore the potential of semi-automated systems to enhance the reliability and efficiency of ONSD measurements, particularly in environments where precision and speed are of paramount importance\cite{fleiderman2023spino,el2023measuring}.

	\subsection{Fully Automated Methods}
	\vspace{2pt}
	Fully automated measurement systems have become increasingly significant in medical imaging due to their potential to reduce operator dependency and increase diagnostic accuracy. A prominent example of this advancement is seen in the development of deep learning-based systems for retinal vessel measurement. Shi et al. demonstrated that these fully automated systems can perform high-throughput image analysis with minimal human intervention, offering a scalable and consistent approach to ocular measurements. This method not only improves the precision of measurements but also significantly enhances the efficiency of clinical assessments, highlighting the broader applicability of fully automated techniques in ophthalmology and beyond\cite{shi2022deep}.
	
	In a related application, Russo et al.explored the use of fully automated methods for measuring intracranial cerebrospinal fluid (CSF) and brain parenchyma volumes in pediatric patients with hydrocephalus. This approach employs advanced segmentation techniques applied to clinical MRI data, enabling accurate and reproducible measurements that are critical for diagnosis and treatment planning. The fully automated system effectively minimizes the variability and time demands associated with manual methods, thereby improving the overall reliability and efficiency of the diagnostic process. These studies collectively underscore the transformative potential of fully automated systems in medical imaging, particularly in contexts that demand high precision and consistency\cite{russo2023fully}.
	
	\begin{figure*}[t]
		\centering
		\includegraphics[width=0.9\linewidth]{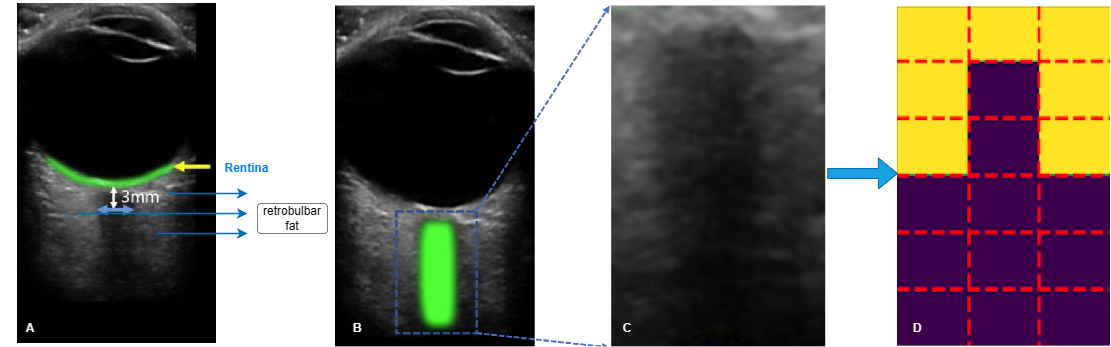}
		\caption{\textbf{Illustration of ONS}: (A), (B), (C) are the visualizations of ONS (D) manual design groud truth of ONS.}
		\label{fig:2}
	\end{figure*}
	
	\section{Methodology}
	\label{sec:method}
	
	\subsection{Dataset}
	Bedside ultrasound examinations were performed on patients and healthy volunteers using a Youkey Q7 device (Wuhan Youkey Biomedical Electronics Co., Ltd., China). B-mode imaging was performed using a linear transducer (6-11 MHz). All patients were admitted to the ICU within 12 hours of traumatic brain injury or cerebrovascular accident and had at least one clinical feature suggestive of increased intracranial pressure (ICP), including changes in consciousness, headache, nausea and vomiting, bradycardia, hypertension, or changes in pupil diameter. The following parameters were recorded for each patient: sex, age, weight, systolic blood pressure (SBP), diastolic blood pressure (DBP), etiology (e.g., ischemic stroke, hemorrhagic stroke), Glasgow Coma Scale (GCS) score, blood glucose level, arterial partial pressure of oxygen (PaO2), arterial partial pressure of carbon dioxide (PaCO2), and pupil diameter,and assess patient status daily.
	
	From February 2021 to November 2021, a total of 21 patients (age: 70 years old, male: 16, female: 5) and 17 healthy volunteers (age: 67 years old, male: 9, female: 9) were enrolled in the Intensive care Unit (ICU) of Minhang Hospital Affiliated to Fudan University. 8) underwent optic nerve sheath diameter (ONSD) ultrasound. For each subject, two ultrasound videos (two for each eye) were acquired for subsequent algorithm analysis. A total of 40 ultrasound videos were included in this study. The ONSD for each video was measured blindly by two sonographers with 5-year experience in using ultrasound. When the frame containing the maximum diameter from the ONSD video was observed, the operator freezed the ultrasound  image and measured the ONSD at a longitudinal depth of 3 mm below the retina. The measurement was performed for 3 times on each video. The ONSD measurement result for the examined video was calculated as the average of the three measurements serving as ground truth for the following comparative study. In this experiment, we will analyze the best keyframes for the acquired ultrasound images using the SLIC super-pixel image segmentation approach. \hyperref[fig:1]{Fig.\ref{fig:1}} describes the keyframe selection mechanism in detail.
	
	This study was approved by the Ethics Committee of Minhang Hospital of Fudan University (Approval No.2022-009-01X) with the informed consent of all participants.
	
	\subsection{Proposed Method}
	Previous reported measurement of ONSD were primarily based on freeze images. Other than earlier methods, our proposed ONSD pipeline automatically provided each video frame with a ONSD imaging quality score, based on which the optimal imaging video frame with the highest score was extracted and used for ONSD measurement. The ONSD automatic measurement system consisted of three steps: (1) Optimal Frame Selection; (2) ONS localization; (3) Weighted boundary refinement.
	
	\subsubsection{Optimal Frame Selection Strategy}
	\added{
		The optimal frame selection strategy comprises three sequential stages (see \hyperref[fig:1]{Fig.\ref{fig:1}-(B,C,D)}):}
	\begin{enumerate}
		\item \added{\textbf{Dynamic Region of Interest (ROI) Localization}: KCF tracking} \citep{zhang2014fast} \added{is applied to each video frame to isolate the optic nerve sheath region, generating a bounding box that dynamically adapts to anatomical motion.}
		
		\item \added{\textbf{Superpixel Segmentation}: Within the KCF-localized ROI, SLIC partitions the image into edge-preserving superpixels, where each cluster is assigned a label based on spatial and intensity coherence.}
		
		\item \textbf{Optimal Frame Scoring}: The overlap between SLIC-generated superpixels and the manually designed ground truth (\hyperref[fig:1]{Fig.\ref{fig:1}-(D)}) is calculated to determine the optimal frame.
	\end{enumerate}
	
	Due to the blurred edges of the optic nerve sheath in ultrasound images, directly segmenting each frame of an ultrasound video to assess image quality is challenging. To address this issue, we approached the problem as a template matching task. Assuming we have a ground truth ONSD, we can calculate a matching score for each frame in the ultrasound video, with the highest score indicating the optimal frame for ONSD measurement. In this context, our proposed optimal frame selection strategy can be divided into two parts: (1) automatically generating segmentation predictions and basic truths of ONSD , and (2) calculating the matching scores and selecting the highest-scoring frame as the optimal one for subsequent ONSD measurement.
	
	\replaced{In ocular ultrasound imaging, retrobulbar adipose tissue surrounding the optic nerve sheath (located approximately 3 mm posterior to the globe) demonstrates hyperechoic characteristics, manifesting as high-intensity pixel regions in the left, right, and superior aspects of the imaging plane (\hyperref[fig:2]{Fig.\ref{fig:2}-(A-C)}). Notably, the optic nerve sheath itself presents as a distinct hypoechoic structure (low pixel intensity), exemplified by the green demarcated region in \hyperref[fig:2]{Fig.\ref{fig:2}-(B)}.}{In ocular ultrasound, the left, right and upper parts of the optic nerve sheath (3mm area in the lower part of the retina) are retrobulbar fat bodies, which exhibit as high echo areas (high pixel values), as shown in \hyperref[fig:2]{Fig.\ref{fig:2}}-(A)(B)(C). Notably, the green area in \hyperref[fig:2]{Fig.\ref{fig:2}}-(B) is a example of optic nerve sheath region. While the optic nerve sheath is a low echo area (low pixel value)}. According to this characteristic, we divided the image into 3$\times$6 regions, and labeled them as $R=[r_1,\cdots,r_{18}]$, where the gray value of pixels in $r_1,r_2,r_3,r_4,r_6,r_7,r_9$ are set to be 255 with the rest being 0 as showed in \replaced{\hyperref[fig:2]{Fig.\ref{fig:2}}-(D)}{\hyperref[fig:2]{Fig.\ref{fig:2}}-(E)}. This is a basic fact of manual design, which can ensure the accuracy of the results based on the doctor's experience for representing ONSD (hyperechoic area). Based on this ground truth, the optimal frame selection strategy will match each video frame with this ground truth to generate scores.
	
	Recall the main idea of our proposed optimal frame selection strategy: for each frame in the video, we computed the overlap area between ONSD ground truth and segmentation result given by Simple Linear Iterative Clustering\cite{achanta2010slic}. The frame whose segmentation result maximally overlapped the manually devised ground truth was selected as the optimal frame. SLIC is an unsupervised segmentation method which has the advantage of preserving the edge information. It segments an image into many regions with its corresponding label as shown on \hyperref[fig:1]{Fig.\ref{fig:1}}-(A). The region with the same color signifies that SLIC classifies the objects into the same categorization. However, it is unknown which subcategories belong to ONSD. \replaced{To distinguish ONSD from the background, we perform label realignment by assigning each superpixel's intensity value as the summation of all pixel values within its corresponding SLIC cluster. This process, visualized in \hyperref[fig:1]{Fig.\ref{fig:1}}-(C), enhances the discriminability between hyperechoic retrobulbar fat and hypoechoic ONSD regions.}{To distinguish ONSD from the background, we substitute each subcategory's pixel value with the sum of all pixel values in that region. The visualization is depicted in \hyperref[fig:1]{Fig.\ref{fig:1}}-(C).} A lighter color indicates a higher pixel value (high echo region) while a darker color implies a smaller pixel value (low echo region). Then, we binarized the k subregions with the highest pixel values as our ONSD segmentation findings as showed on \hyperref[fig:1]{Fig.\ref{fig:1}}-(E).
	
	After obtaining the segmentation results for each frame, we further calculated the overlap area between manually designed ground truth ONSD and the generated ONSD segmentation by:
	\begin{equation}
		S = \frac{2 \left| Y \cap \hat{Y} \right|}{\left| Y \right| + \left| \hat{Y} \right|}
	\end{equation}
	where $Y$ and $\hat{Y}$ are the manually designed ground truth ONSD and the generated ONSD segmentation for each frame, respectively. We then derived a collection of ONS quality score with the optimal frame we selected to measure ONSD as 
	\begin{equation}
		\arg\max_{i} \frac{2 \left| Y \cap \hat{Y_i} \right|}{\left| Y \right| + \left| \hat{Y_i} \right|}
		\quad \text{for } i = 1, 2, \dots, 18
	\end{equation}
	
	\begin{figure*}[t]
		\centering
		\includegraphics[width=0.9\linewidth]{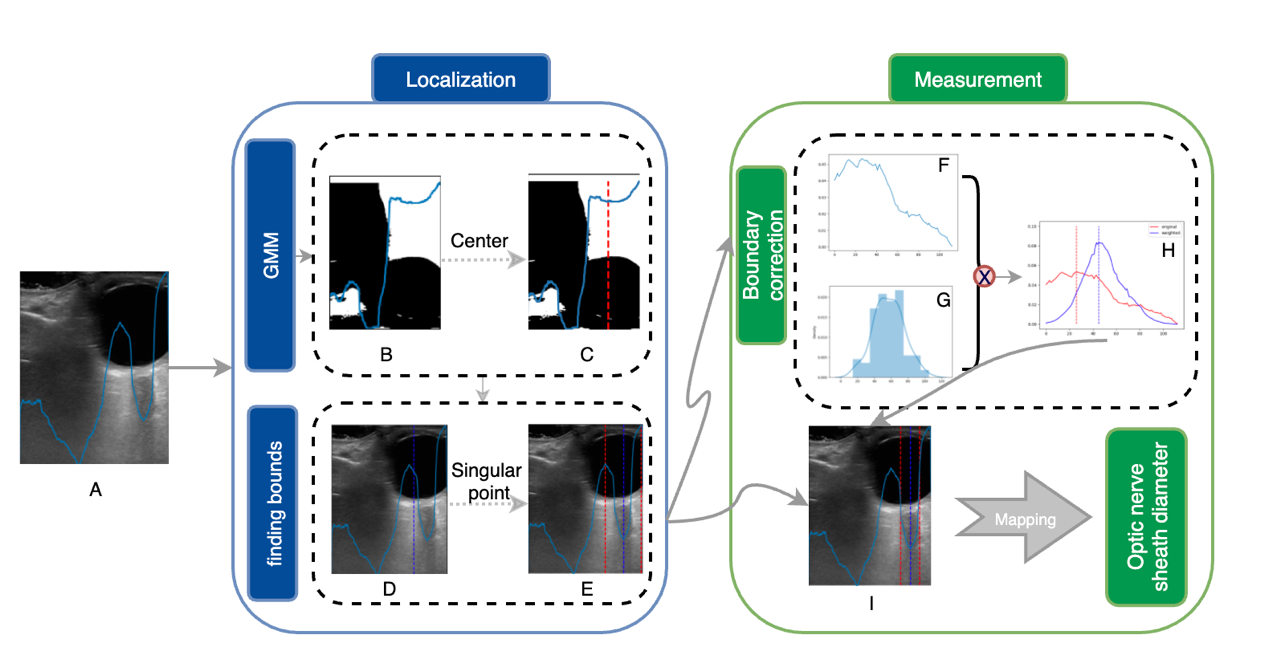}
		\caption{\textbf{Overview of the proposed measurement method}: (A) original image with signal $v(n)$; (B) Foreground extraction by GMM; (C) Center line of Foreground segmentation Center; (D)Center line of Foreground segmentation Center on original image; (E) Localizing result; (F) KL divergence change from left boundary to center; (G) Normal distribution; (H) weighted result; and (I) Optic nerve sheath measurement results}
		\label{fig:3}
	\end{figure*}
	
	After obtaining the optimal frame, we thenceforth performed our proposed ONSD measurement method. \replaced{Prior to SLIC segmentation, the KCF tracker initializes a ROI by exploiting temporal coherence across consecutive frames. This spatial prior constrains subsequent processing to anatomically relevant regions, effectively suppressing extraneous tissue interference.}{Notably during the stage of identifying optimalframe, we analysedanalyzed the part of image given by KCF tracking algorithm to select optimal frame.} 
	
	The overview of our proposed measurement method is presented on \hyperref[fig:3]{Fig.\ref{fig:3}}. It was divided into two stages: localization and measurement. The stage of localization entailed estimating the approximate position of the optic nerve sheath in preparation for the following measurement. In the stage of measurement, a boundary adjustment was utilized to get the diameter of optic nerve sheath based on the stage of location.
	
	\subsubsection{Optic Nerve Sheath Localization}
	Previous studies have shown that the left and right boundaries of ONSD can be detected by calculating the sum of the pixel values of each column and each row respectively\cite{moore2020automatic}. Assuming that the original image is a $N \times M$ 2-D array, the pixel value on $(n,m)$ is denoted as $P(n,m)$. A one-dimensional signal can be calculated as followed: 
	
	\begin{equation}
		v(n) = \sum_{m=1}^{M} P(n, m) \quad \text{for } n = 1, \cdots, N
	\end{equation}
	
	\begin{figure}[h]
		\centering
		\includegraphics[width=0.9\linewidth]{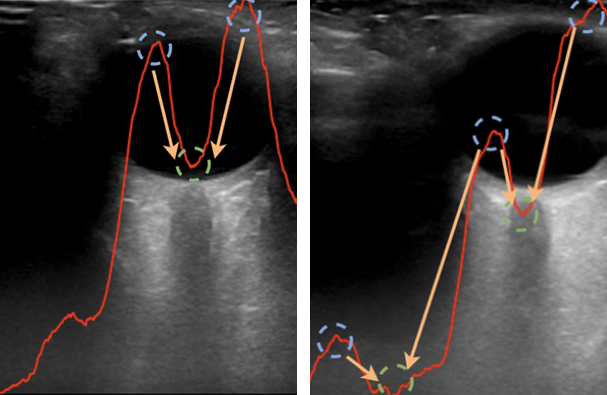}
		\caption{Left: Good sigmal $v(n)$. Right: Bad signal $v(n)$}
		\label{fig:4}
	\end{figure}
	
	\replaced{These $v(n)$ are shown in \hyperref[fig:3]{Fig.\ref{fig:3}-(A)} and \hyperref[fig:4]{Fig.\ref{fig:4}}, where the optic nerve sheath center corresponds to a local minimum flanked by two local maxima.}{The signals $v(n)$ is showed on \hyperref[fig:3]{Fig.\ref{fig:3}}-(A).} Notably, using the whole image to calculate the signal $v(n)$, as opposed to a part of the image given by KCF tracking algorithm, accentuates the contrast between pixel values and improves the measurement’s precision\cite{soroushmehr2019automated}. Hence, the follow calculation of signal $v(n)$ is based on the whole image. An important property of this signal is the center of optic nerve sheath located on the trough between two peaks as shown on \hyperref[fig:4]{Fig.\ref{fig:4}-(left)} . We can use this property of $v(n)$ to locate optic nerve sheath. Hence the problem is changed to find $n=i$ given a signal $v(n)$, that satisfies:
	
	\begin{equation}
		v(n) =
		\begin{cases}
			\text{local minimum}, & x = i \\
			\text{exist a local maximum}, & x > i \\
			\text{exist a local maximum}, & x < i
		\end{cases}
	\end{equation}
	
	However, previous methods based on this signal $v(n)$ to measure the diameter of optic nerve sheath did not consider real situations where the data may be complex\cite{meiburger2020automatic} due to the noise introduced in ultrasound imaging. For example, in the signal $v(n)$ as shown in \hyperref[fig:4]{Fig.\ref{fig:4}-(right)}, there exists three peaks and two troughs, while only one indicates the center of the optic nerve sheath. In this scenario, there is no unique $n=i$ that statisfies the equation \textit{x}. To tackle this problem, instead directly took derivative on signal $v(n)$ and randomly searched singular points, we proposed a region initialization search strategy to obtain the starting point. Based on this starting point, using the idea of gradient descent method, we can quickly and accurately locate the center of the optic nerve sheath. Starting from the determined center, we subsequently searched the rough localization left and right boundary around the optic nerve sheath by taking the derivative of signal $v$.
	
	\subsubsection{Region Initialization Search Strategy}
	The optic nerve and optic nerve sheath, both part of nerve tissue, are surrounded by retrobulbar fat \cite{hansen1996subarachnoid}. These structures present distinct characteristics on ultrasound images. Specifically, in terms of pixel distribution, the optic nerve and optic nerve sheath exhibit low gray values, indicating low echogenicity, while the retrobulbar fat shows high gray values, with the eyeball and surrounding areas appearing black. As illustrated in \hyperref[fig:2]{Fig.\ref{fig:2}}-(D), this characteristic allows for the segmentation of foreground and background to distinguish the retrobulbar fat area from other regions.  For this purpose, we employed the Gaussian Mixture Model (GMM). This well-established clustering algorithm uses the Gaussian probability density function to cluster data into several components. The binary segmentation mask provided by the GMM, denoted as $M(n,m)$, contains values of either 255 or 0 for each position $(n,m)$, as described by \cite{he2010laplacian}. The approximate segmentation result is shown in \hyperref[fig:3]{Fig.\ref{fig:3}}-(B). Consequently, the signal $k(n)$ is derived based on this segmentation mask:
	\begin{equation}
		k(n) = \frac{\sum_{m=1}^{M} M(n, m)}{\sum_{n=1}^{N} \sum_{m=1}^{M} M(n, m)} = 1
	\end{equation}
	
	The optimal initial search point $d$ is the center of signal $k(n)$ for the purpose of efficient searching. It can be calculated by solving $k(n)=0.5$. Then the center of the optic nerve sheath is positioned at a through in signal $V(n)$, with two peaks on the left and right sides. \replaced{The core objective of this analytical phase is to identify the characteristic trough in the signal profile that uniquely corresponds to the optic nerve sheath center, distinguished by being flanked symmetrically by two local maxima.}{The goal of our purpose is to find this unique through.} We calculated the current gradient for signal $V$ assuming our starting localization point is $n=d$ and the center of the optic nerve sheath being searched is $d_{center}$. Notably, the gradient for $n=d$ is calculated by $\Delta V(n=d)=V(d)-V(d-1)$. The search proceeds to the left if $\Delta V(n=d)$ is negative, and vice versa. Following is the pseudocode for the entire search procedure:
	
	\begin{algorithm}
		\caption{Locating Center}
		\begin{algorithmic}[1]
			\REQUIRE signal $v(n)$, signal $k(n)$, $d$ where $k(d)=0.5$
			\ENSURE output $d$
			\WHILE{$\Delta v(n) \neq 0$}
			\IF{$\Delta v(n) < 0$}
			\STATE $d=d+1$
			\ENDIF
			\IF{$\Delta v(n) > 0$}
			\STATE $d=d-1$
			\ENDIF
			\ENDWHILE
		\end{algorithmic}
	\end{algorithm}
	
	After locating and determining the center $d_{center}$, easily $d_{left}$ and $d_{right}$ are the first peak on the left and right for $v(n)$, which is described in \hyperref[fig:3]{Fig.\ref{fig:3}}-(E)
	
	\subsubsection{Weighted boundary refinement method}
	The localization result of the preceding phase was coarse and larger than the actual optic nerve sheath region. The uncertainty surrounding the edge of the optic nerve sheath presents a challenge to the detection of the ONS. We presented a KL-divergence-based approach for precise detection of the optic nerve sheath's boundary\cite{zhang2017improved}. KL divergence can be utilized to quantify the degree of dissimilarity between two distributions. Consider two discrete probability distributions, $GL_d$ and $GL_{\text{center}}$. $Q$ represents the grayscale distribution of pixels in the original image, where the x-coordinate is $d_{\text{center}}$ determined by the region initialization search approach. $GL_d$ represents the gray distribution of pixels in the original image whose x-coordinates are located in $\left[ d_{\text{left}}, d_{\text{center}} \right]$. Then the KL divergence between $d_{\text{left}}$ and $d_{\text{center}}$ is given by:
	
	\begin{equation}
		D_{KL}(GL_{d_{\text{left}}} \parallel GL_{\text{center}}) = - \sum_{x} GL_{d_{\text{left}}}(x) \log \frac{GL_{\text{center}}(x)}{GL_{d_{\text{left}}}(x)}
	\end{equation}
	
	Define the left position KL divergence signal: 
	
	\begin{equation}
		L(d) = D_{KL}(GL_{d} \parallel GL_{\text{center}})
	\end{equation}
	
	\hyperref[fig:3]{Fig.\ref{fig:3}}-(F) depicts the entire signal $L(d)$, where $d_{\text{left}} \leq d \leq d_{\text{center}}$, which denotes the similarity transition from the left bound to the center of the optic nerve sheath. An edge or boundary intuitively defines a change between two items, such as the optic nerve sheath. It signifies the shift from one state to another when it is crossed. Consequently, the unique point of signal $L(d)$ is most likely the optic nerve sheath's edge. However, there are numerous uncertain singularity points. Directly using $L(d)$ to determine the exact optic nerve sheath boundary in this circumstance may be erroneous. \deleted{To reduce measurement error, we generated weights based on positions from a normal distribution $N \sim (\mu, \sigma^2)$, with $\mu = \frac{d_{\text{center}} - d_{\text{left}}}{2}$ and $\sigma = \frac{d_{\text{center}} - d_{\text{left}}}{6}$,  and multiple it to $L(d)$. The modified position KL divergence signal $\tilde{L}(d)$ is:}
	
	\added{To address the boundary uncertainty inherent in our ultrasound imaging, we utilize a weighted KL-divergence method grounded in clinical prior knowledge where the KL divergence is adopted to quantify the dissimilarity between grayscale distributions across candidate boundaries. Unlike gradient-based methods, KL-divergence directly models probabilistic transitions between tissue types, providing robustness against speckle noise common in ultrasound (\hyperref[fig:3]{Fig.\ref{fig:3}-(F)}). The probability density function of normal distribution is given by:}
	
	\begin{equation}
		\tilde{L}(d) = \frac{1}{\sigma \sqrt{2\pi}} e^{-\frac{1}{2} \left(\frac{d - \mu}{\sigma}\right)^2} L(d)
	\end{equation}
	
	\begin{figure*}
		\centering
		\includegraphics[width=0.8\linewidth]{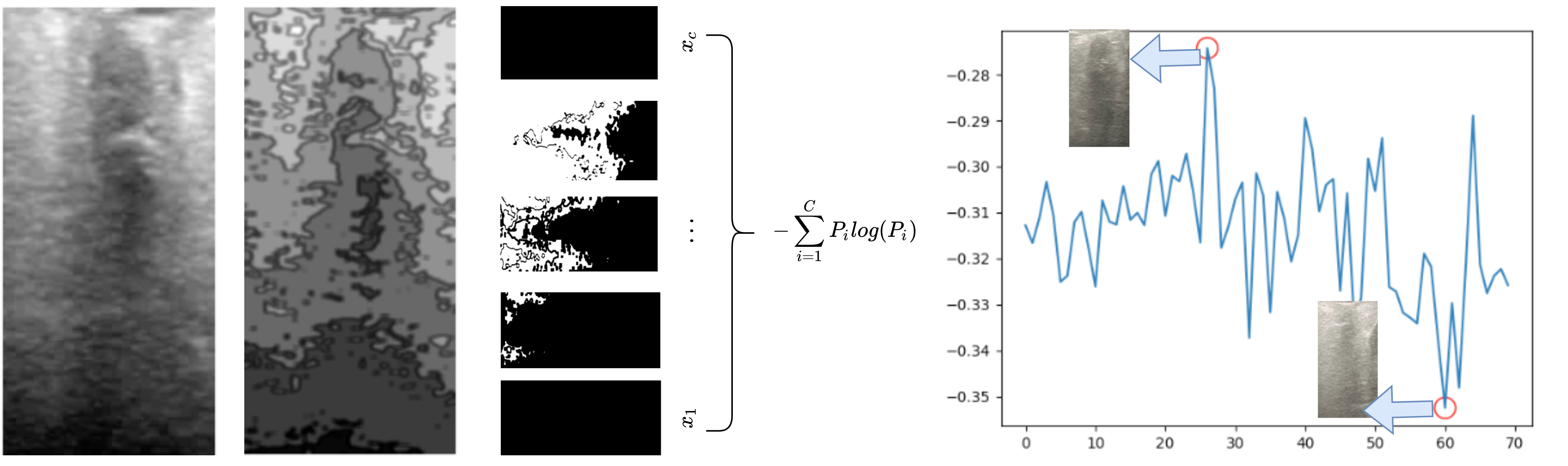}
		\caption{Capturing key frames from dynamic video data automatically. Extracting two peaks based on information entropy.}
		\label{fig:5}
	\end{figure*}
	
	\begin{figure*}
		\centering
		\includegraphics[width=1.0\linewidth]{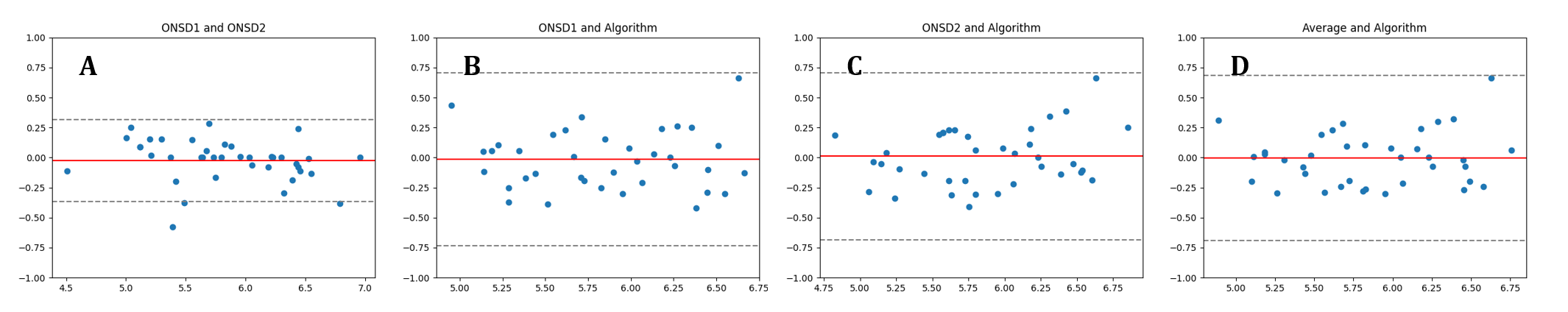}
		\caption{\textbf{Bland-Altman plot.} Comparison of different methods.(A) Two experts' manual measurements. (B) $ONSD_1$ and proposed method. (C) $ONSD_2$ and proposed method. (D) Averaged result of two experts' and proposed method.}
		\label{fig:6}
	\end{figure*}
	\added{where the weighting function parameters $(\mu,\sigma)$ are anatomically derived: (1) $\mu=\frac{d_{\text{center}}-d_{\text{left}}}{2}$ positions the prior at the geometric center of the KCF-localized ROI, aligning with the sheath's expected bilateral symmetry. (2) $\sigma = \frac{d_{\text{center}} - d_{\text{left}}}{6}$ confines 99.7\% of the distribution within the initial search region ($d_{\text{left}}\pm 3\sigma$), preventing over-reach into physiologically implausible areas.}
	
	$\tilde{L}(d)$ is shown on \hyperref[fig:3]{Fig.\ref{fig:3}}-(H) (blue line) while $L(d)$ is showed on  \hyperref[fig:3]{Fig.\ref{fig:3}}-(H) (red line). Define the precise left boundary of the optic nerve sheath as $\tilde{d}_{\text{left}}$, which satisfies $\Delta \tilde{L}(\tilde{d}_{\text{left}})=0$. Since when the gradient is equal to zero, the change in similarity from the left boundary to the center of the optical nerve sheath begins to transition from one state (dissimilar) to another (similar). $\tilde{d}_{\text{left}}$ is an accurate boundary of our optic nerve sheath. Similarly, we can derive $\tilde{d}_{\text{right}}$. The refinement result is shown on \hyperref[fig:3]{Fig.\ref{fig:3}}-(I). The final actual diameter of optic nerve sheath is measured by:
	
	\begin{equation}
		D=\text{Mapping System}(\tilde{d}_{\text{right}}-\tilde{d}_{\text{left}})
	\end{equation}
	The mapping system depends on the acquiring machine and the parameters settings.
	
	Based on the above methods, we develop an automatic measurement ONSD system to automatically collect and measure key frames in video data. The system operates on information entropy. Information entropy is used to measure how much information is contained in the data and is given by the following formula:
	
	\begin{equation}
		IC=-\sum_{i=1}^{C} P_i\log P_i
	\end{equation}
	where $P_i$ is the probability of each case. \hyperref[fig:5]{Fig.\ref{fig:5}} demonstrates the process we extract key information based on information entropy (peak values).
	
	\added{Through our method and parameterization, we ensure that boundary detection respects both spatial coherence (by SLIC superpixels) and clinical plausibility (by anatomical constraints). The validation on our videos also showed $\le 2\%$ deviation from expert-drawn boundaries when $\sigma$ ranged between 15-20\% of ROI width, confirming the parameter stability.}
	
	\subsection{Evaluation Matrix}
	To demonstrate the efficacy of our proposed method, we calculated the average error between our proposed method and two experts' manual measurements via
	
	\begin{equation}
		e = \frac{1}{n} \sum_{k=1}^{n} \left| 1 - \frac{\text{ONSD}_1(k)}{\text{ONSD}_2(k)} \right| \times 100
	\end{equation}
	Where $n$, $ONSD_1$ and $ONSD_2$ are the number of ultrasound images , automated measurement using our proposed method, and the average manual measurement results by two experts, respectively. The mean square error is used to confirm the discrepancy in measurements between two experts:
	
	\begin{equation}
		MSE = \frac{1}{n_u} \|ONSD_1 - ONSD_2\|_2
	\end{equation}
	
	Furthermore, we determined the correlation coefficient and 95\% confidence interval between the ground truth diameter and the automated measured diameter.
	
	\section{Empirical Results}
	
	\begin{figure}[h]
		\centering
		\includegraphics[width=1.0\linewidth]{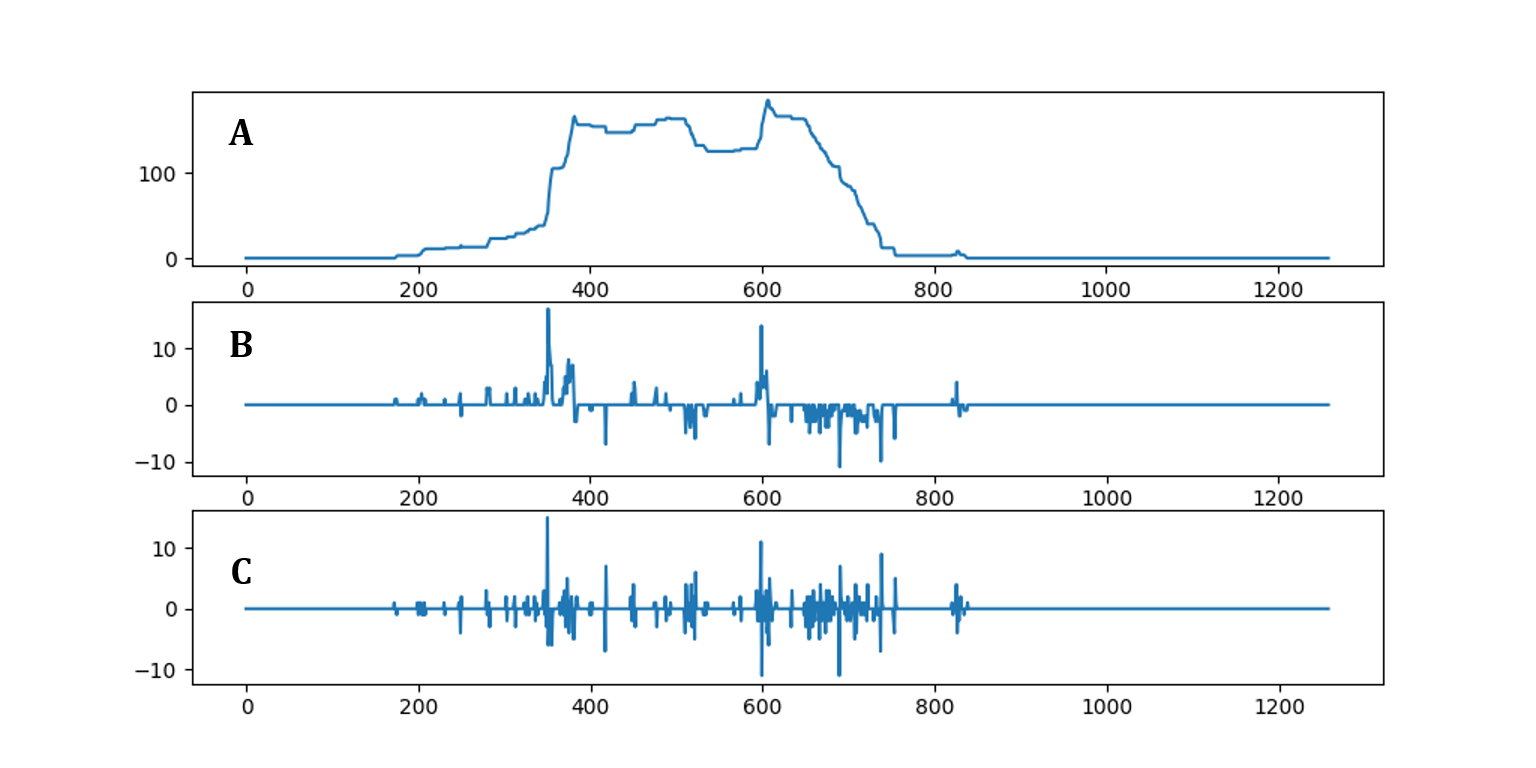}
		\caption{\textbf{Processing of ONSD figure}. (A) Smoothing. (B) First-order Difference. (C) Second Difference.}
		\label{fig:7}
	\end{figure}
	
	Our study smoothed and differentiated the extracted optic nerve sheath image to eliminate noise and detect edges, which is convenient for automatic measurement of ONSD. \hyperref[fig:7]{Fig.\ref{fig:7}}-(A) shows the smoothed optic nerve sheath data, and \hyperref[fig:7]{Fig.\ref{fig:7}}-(B) and (C) are the curves obtained by first-order and second-order differentials of the smoothed data, respectively. Such processing can effectively improve the edge detection accuracy of our proposed method.
	
\begin{figure}
		\centering
		\includegraphics[width=0.8\linewidth]{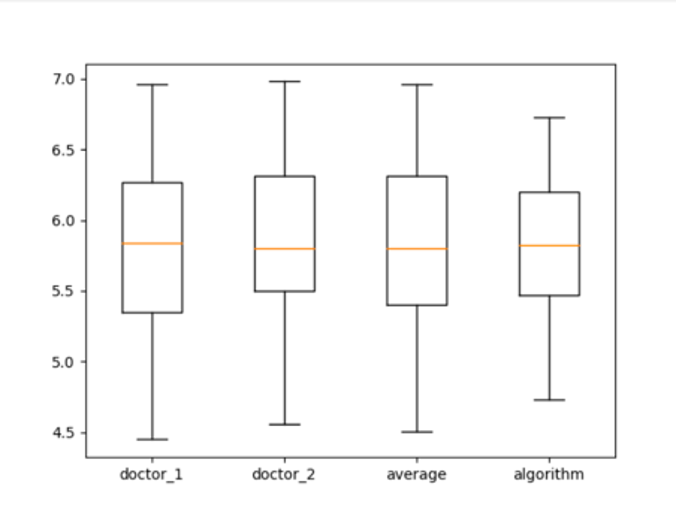}
		\caption{\textbf{Box plot of four methods of measurements.}}
		\label{fig:8}
	\end{figure}

\begin{figure}[h]
		\centering
		\includegraphics[width=1.0\linewidth]{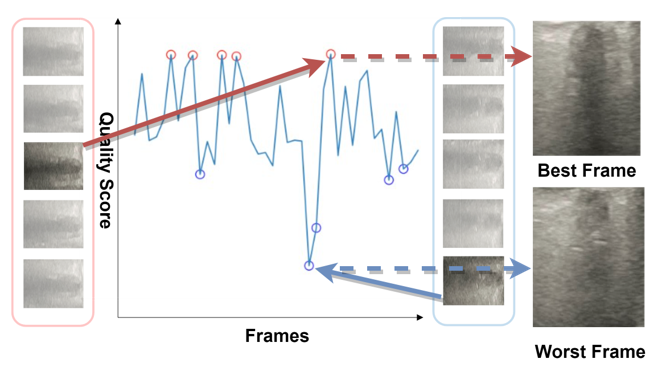}
		\caption{Visualization result of our proposed optimal frame selection method.}\label{fig:9}
	\end{figure}

	To demonstrate the superiority of our method, our study used Bland-Altman plot and box plot to compare the manual measurements of experts with our automatic measurement method. \hyperref[fig:6]{Fig.\ref{fig:6}} and \hyperref[fig:8]{Fig.\ref{fig:8}} show the comparison between the measurements of two doctors, our proposed method and the two doctors, and the mean of the two doctors' measurements and our method. Obviously, despite the large discreteness, our algorithm does not show inconsistency with the manual measurements of experts. All data points are within the limits of agreement (LoA) without obvious systematic errors. Although the overall measurement trend of the algorithm is consistent with that of the doctors, it shows greater discreteness and a slightly lower median in a smaller range of measurement values. Further algorithm optimization, especially in the case of low measurement values, can improve consistency and accuracy. Based on the Qualify Score calculated by our method and the comparison results of the above measurements, we can obtain the optimal frame selection result (as shown in \hyperref[fig:9]{Fig.\ref{fig:9}}).
	
	\begin{table*}[h!]
		\centering
		\caption{\replaced{Comparison of ONSD measurements (unit: mm) between proposed method and expert annotations}{Comparison of ONSD measurements and Proposed Method.}}
		\begin{tabular}{l c c c c c }
			\hline
			\multirow{2}{*}{\textbf{Analysis}} & \multirow{2}{*}{\textbf{Mean Error}} & \multirow{2}{*}{\textbf{MSE}} & \multirow{2}{*}{\textbf{ICC}} & \multicolumn{2}{c}{\textbf{95\% Confidence Interval}} \\ \cline{5-6} 
			&  &  &  & \textbf{Lower limit} & \textbf{Upper limit} \\ \hline
			\textit{ONSD$_1$ versus ONSD$_2$} & 0.020 & 0.027 & 0.952 & 0.911 & 0.974 \\ \hline
			\textit{ONSD$_2$ versus Proposed Method} & 0.043 & 0.055 & 0.787 & 0.625 & 0.88 \\ \hline
			\textit{ONSD$_1$ versus Proposed Method} & 0.044 & 0.057 & 0.754 & 0.573 & 0.86 \\ \hline
			\textit{Average versus Proposed Method} & 0.040 & 0.054 & 0.782 & 0.616 & 0.877 \\ \hline
		\end{tabular}
		\label{tab:1}
	\end{table*}
	
	\hyperref[tab:1]{Tab.\ref{tab:1}} \replaced{summarizes the mean errors (in millimeters, mm) and intraclass correlation coefficients (ICC) for all comparison groups.}{depicted the results of mean errors of our methods.} Specifically, our proposed method gave mean error of 0.044 and 0.043 compared with $ONSD_1$ and $ONSD_2$, respectively. In addition, we calculated MSE value to evaluate the difference between two experts' measurements. The MSE between two experts' measurement is 0.027(mm). The intraclass correlation coefficient (ICC) between two experts' measurement is 0.952, while the one between the algorithm and two experts' average measurement is 0.782. 
	
	The results showed that there is no significant difference and variability between the proposed method in terms of ONSD and manual measurements (\hyperref[tab:1]{Tab.\ref{tab:1}}). This indicates that our proposed algorithm is highly reliable in measurement, \replaced{with an ICC greater than 0.75 and a mean error less than 0.05 mm compared to two experts.}{with an ICC greater than 0.75 and an average error less than 0.05 made by two experts.}

	\section{Discussion}
	In previous studies, although several methods have been proposed for ONSD localization and measurement, each of these methods has limitations. The existing ONSD measurement method is mainly applied to image-level data manually selected by experts from dynamic ultrasound videos, which greatly relies on the subjective judgment of experts. In contrast, our study presents for the first time an automated measurement method based on ultrasound video that significantly reduces the reliance on expert judgment. Our method is divided into two main stages: selecting the best frames from the ultrasound video and then performing ONSD measurements based on the selected best frames. We designed a detailed benchmark to represent the optimal optic nerve sheath and evaluated the quality of each frame by calculating the overlap between the manually designed benchmark and the segmentation results provided by the SLIC hyperpixel method to determine the optimal frames for further ONSD measurements. Subsequently, we used a boundary adjustment method to accurately detect the boundaries of the optic nerve sheath.
	
	To address these issues, we propose an ONSD measurement system that can automatically filter the best frames from video data. We first introduce a well-designed benchmark, and then evaluate the quality of the frames by calculating the overlap between the manually designed benchmark and the segmentation results given by the SLIC method to determine the best frames. Based on the selected optimal frames, we propose a weighted fine-grained measurement method to compute the ONSD. \added{To mitigate the potential subjectivity introduced by the manually designed echogenicity-based benchmark, we performed a comparative validation with two experienced critical care clinicians. Each independently selected optimal frames from 20 randomly sampled ultrasound videos. These expert-selected frames were processed through our automated ONSD measurement pipeline, and the results were compared to those derived from our automatic frame selection strategy. The mean absolute difference in measured ONSD was 0.037 mm (95\% CI: 0.015–0.059 mm), with an intraclass correlation coefficient (ICC) of 0.801, demonstrating high consistency without significant bias.}
	
	\added{Our findings align with previous studies comparing automated and clinician-selected approaches in ONSD analysis. For instance, \cite{rajajee2021novel} and \cite{soroushmehr2019automated} reported strong agreement between automated segmentation methods and expert measurements. \cite{moore2020automatic} further validated the clinical viability of automated frame selection in point-of-care ultrasound. These results collectively support the reliability and clinical interpretability of our frame scoring strategy, despite its manual benchmark initialization.}
	
	\replaced{This study demonstrates promising performance under a standardized single-center protocol; however, the limited sample size (40 videos) and lack of cross-center data may constrain generalizability. Real-world deployment faces challenges such as variability in ultrasound systems (e.g., GE Logiq and Philips EPIQ), operator-dependent measurement bias (e.g., probe angle deviations leading to ~0.12 mm ONSD shift), and anatomical diversity across patient populations. To evaluate robustness, we introduced synthetic noise and motion artifacts (SNR = 15 dB), observing MAE between 0.04 mm and 0.06 mm. On an external dataset (ONSD-Benchmark, n = 32), the model achieved an ICC of 0.74, comparable to the internal ICC of 0.78. These findings suggest reasonable cross-domain reliability, though further validation on multi-center cohorts is needed.}{However, there are some limitations in this study. First, in the measurement stage we used a normal distribution weighting method to adjust the boundaries, which is a relatively strong a priori assumption.  Due to the low signal-to-noise ratio, low contrast, and blurred boundaries of ultrasound images, our a prior assumption may affect the performance of the system, although our experimental results have demonstrated the effectiveness of the algorithm. Second, our method was tested only on video data acquired at a single center. Different ultrasound devices and different operating methods may lead to differences in video quality, which may affect the generalizability and accuracy of our method.}
	
	\added{Although convolutional neural network (CNN)-based approaches\cite{xiao2024automatic} can achieve high measurement accuracy under ideal conditions, their reliance on large-scale expert-labeled datasets (typically 300–500 annotated videos or images) and high-performance computing hardware limits their applicability in resource-constrained settings, such as rural clinics or public health emergency sites. For instance, in underserved regions, less than 5\% of healthcare facilities are equipped to implement deep learning pipelines. In contrast, our method integrates anatomical priors with lightweight image processing, achieving clinically acceptable accuracy (ICC > 0.78) under zero-annotation conditions. Furthermore, emerging regulatory frameworks for medical AI (EU MDR 2023 Annex II) emphasize the need for decision traceability, posing compliance challenges for black-box CNN solutions. By employing explicit superpixel-based relabeling and KL-divergence-driven boundary refinement, our approach offers interpretable decision logic better aligned with clinical trust requirements. In high-resource tertiary hospitals, future work may explore integrating our anatomical constraint framework with CNN-based feature extractors to balance interpretability, efficiency, and accuracy.}
	
	In future studies, we plan to validate the performance of our method on datasets from different centers to ensure its robustness and accuracy under different equipment and operating conditions. In addition, we intend to introduce some cutting-edge image processing methods, such as deep learning techniques, to further improve the accuracy and efficiency of the measurements.
	
	\section{Conclusion \added{and Future Work}}
	In this study, we proposed a novel \added{explainable} computer aided method to automatically identify the optimal frame in the video and complete the ONS localization and ONSD measurement. We used mean error, mean square error (MSE), intraclass correlation coefficient (ICC) and Cohen's d as evaluation index to compare the measurement results from our proposed method and two experts. The results showed that the proposed method can accurately measure ONSD, with the potential of improving doctors' work efficiency to a certain extent. \added{Our approach not only demonstrates the potential for application in ONSD measurements, but also provides a new idea for automated ophthalmic ultrasound image analysis. By reducing the reliance on manual intervention and expert judgment, our method is expected to improve diagnostic efficiency and consistency in clinical practice.} In the future, we plan to perform further validation of our algorithm on data sets from multi-centers. \added{We will also incorporate acoustic-model-based data synthesis and semi-supervised learning to improve generalizability.}

	\deleted{Our approach not only demonstrates the potential for application in ONSD measurements, but also provides a new idea for automated ophthalmic ultrasound image analysis. By reducing the reliance on manual intervention and expert judgment, our method is expected to improve diagnostic efficiency and consistency in clinical practice.}
	
	\section*{Acknowledgement}
	\replaced{The authors wish to thank the anonymous referees for their thoughtful comments, which helped in the improvement of the presentation.}{This work was partially supported by the National Natural Science Foundation of China (Grant No. 82151318); Science and Technology Commission of Shanghai (Grant No. 22DZ2229004, 22JC1403603, 21Y11902500);the Key Research \& Development Project of Zhejiang Province (2024C03240 ); Scientific Development funds for Local Region from the Chinese Government in 2023 (Grant No. XZ202301YD0032C); Jilin Province science and technology development plan project (Grant No. 20230204094YY), 2022 "Chunhui Plan" cooperative scientific research project of the Ministry of Education.}
	
	\section*{Declaration of Competing Interests}
	The authors declare that they have no known competing financial interests or personal relationships that could have appeared to influence the work reported in this paper.
	
	\section*{\added{Funding Source}}
	\addcontentsline{toc}{section}{New Section}
	
	\added{
		This work was partially supported by the National Natural Science Foundation of China (Grant No. 82151318); Science and Technology Commission of Shanghai (Grant No. 22DZ2229004, 22JC1403603, 21Y11902500);the Key Research \& Development Project of Zhejiang Province (2024C03240 ); Scientific Development funds for Local Region from the Chinese Government in 2023 (Grant No. XZ202301YD0032C); Jilin Province science and technology development plan project (Grant No. 20230204094YY), 2022 "Chunhui Plan" cooperative scientific research project of the Ministry of Education.
	}
	
	\section*{Authors Contributions}
	\textbf{Renxing Li}:Methodology, Software, Validation, Writing-review \& editing. \textbf{Weiyi Tang}:Conceptualization, Data curation, Validation. \textbf{Peiqi Li}:Methodology, Project administration, Writing-original draft, review \& editing. \textbf{Qiming Huang}:Methodology, Project administration, Visualization, Writing-original draft. \textbf{Jiayuan She}:Methodology, Visualization, Writing-review \& editing. \textbf{Shengkai Li}:Conceptualization, Project administration. \textbf{Haoran Xu, Yeyun Wan, Jing Liu}:Project administration. \textbf{Jiangang Chen, Xiang Li, Hailong Fu}: Conceptualization, Supervision, Methodology.
	
	\section*{Ethics Approval and Consent to Participate}
	This study was approved by the Ethics Committee of Minhang Hospital of Fudan University (Approval No.2022-009-01X) with the informed consent of all participants.
	
	\section*{Data Availability}
	The authors do not have the permission to share the data.
	
	\section*{Code Availability}
	The datasets used and/or analyzed during the current study are available from the corresponding authors on reasonable request.

	\bibliographystyle{elsarticle-num-names}
	\bibliography{Reference.bib}

\end{document}